# Integrating machine learning concepts into undergraduate classes

Chinmay Sahu, Blaine Ayotte, Mahesh K. Banavar
Department of Electrical and Computer Engineering
Clarkson University
Potsdam, NY 13699.
{sahuc, ayottebj, mbanavar}@clarkson.edu

*Abstract*—In this innovative practice work-in-progress paper, we compare two different methods to teach machine learning concepts to undergraduate students in Electrical Engineering. While machine learning is now being offered as a senior-level elective in several curricula, this does not mean all students are exposed to it. Exposure to the concepts and practical applications of machine learning will assist in the creation of a workforce ready to tackle problems related to machine learning, currently a "hot topic" in industry. To this end, the authors are working on introducing Electrical Engineering students to machine learning in a required, Junior-level, Signals and Systems course.

The main challenge with teaching machine learning in a junior-level course involves requiring students to appreciate the linkages between complex concepts in linear algebra, statistics, and optimization. While it can be argued that Junior-level students should have seen concepts in some of these topics, requiring them to apply these topics together is a challenge. Therefore, in order to assist students to better grasp these concepts, we provide them with hands-on activities, since immersive experiences will help students appreciate the practical uses of machine learning.

In a previous approach, authors held stand-alone workshops where students in the class were given Android apps for data collection, followed by different sets of hands-on activities. While this approach showed promise, several students indicated that the stand-alone workshop lacked context. To alleviate these concerns, in the Fall semester of 2020, the authors tried a different approach. Students were provided hands-on activities side-by-side with regular course content enabling links to be made with machine learning throughout the course and providing better context to the content being presented.

Preliminary assessments indicate that this approach promotes student learning. While students prefer the proposed side-by-side teaching approach, numerical comparisons show that the workshop approach may be more effective for student learning, indicating that further work in this area is required.

Keywords— STEM, machine learning, signal processing, projects, hands-on activities

## I. Introduction

Machine learning (ML) and artificial intelligence (AI) are quickly becoming mainstream technologies in the industry [1-3], and students are becoming aware of the possibilities of these technologies [4-7], leading to a need to teach these concepts at the undergraduate level. However, consider that to teach machine learning involves introducing students to complex concepts in statistics, optimization, linear algebra, and decision theory, in which undergraduate students generally do not have appropriate prerequisites. Introducing these complex topics without proper care can lead students to lose enthusiasm for this very important topic.

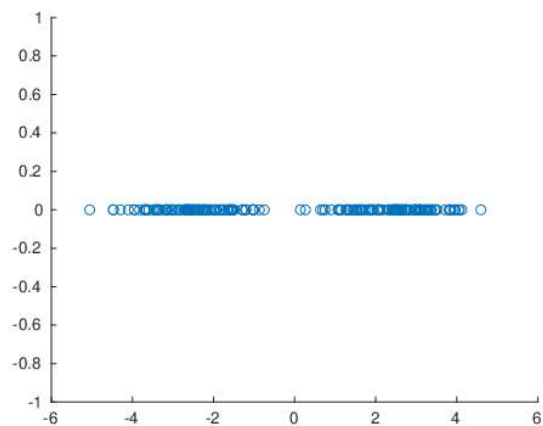

Figure 1. Example data provided to students demonstrating the application of machine learning for classification.

To help with introducing ML to undergraduate students, prior work [4, 5] has focused on content delivery through stand-alone workshops. Such methods have been shown to be promising. Students who participate in the workshop have shown evidence of learning and are generally positive about their experiences in the workshop. However, many of the students also express that such workshops lack context, and links to the main topics in the course are not made.

This feedback led us to the investigation in this paper. If students like the content and activities in a workshop, but believe context is lacking, is there a way to introduce these concepts side-by-side in a course? To this end, we tried a different approach. In the fall semester of 2020, instead of stand-alone workshops, we delivered material synchronized with course content. This allowed students to see the machine learning concepts in the context of the material covered in the course and draw parallels as appropriate. The main advantage of our approach is that students see the context and relevance as part of their course. The challenge with this approach is to integrate the ML content in such a way that it fits naturally with the course material being covered. This type of coverage, naturally, also includes the discussion of both ML and course topics, side-by-side, in class, to improve student understanding.

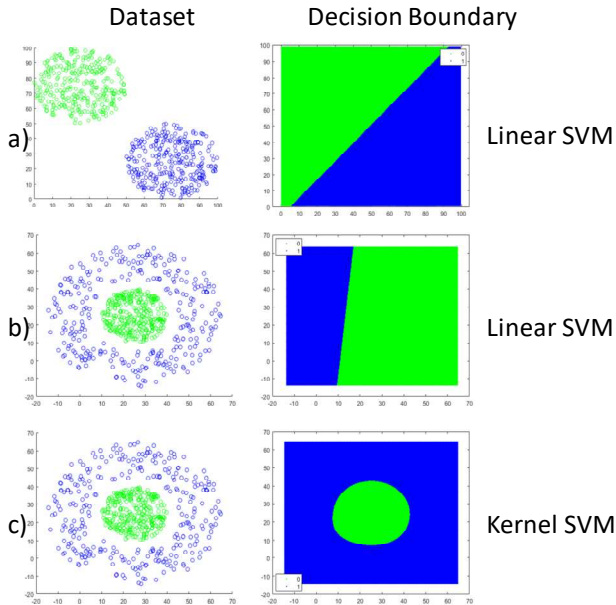

Figure 2. Example of the datasets used in the stand-alone workshop. The two different datasets shown on the left can be classified with high accuracy. Using such visualizations, students learn that depending on the data distribution, the algorithm that provides the best results can change.

To trial our approach, we selected the Junior-level Signals and Systems course at our university, which introduces students to concepts such as orthogonality, dimensionality, and gives them a first look at manipulating signals. Concepts such as principal component analysis can be introduced via orthogonality and dimensionality, and easily linked to concepts in the course such as the Fourier series. These concepts were delivered with projects side-by-side with class concepts. For example, we used principal component analysis (PCA) alongside the Fourier series, both using concepts from dimensionality and orthogonality. Other activities included dimensionality reduction, a communications example with classification and noise analysis, all linked to signal processing concepts. An additional advantage of using these examples is the exposure to students of applications of probability and statistics.

The effectiveness of this approach was evaluated using a pre-/post-test and a survey. Students showed improvement in the areas they were exposed to via the projects, and expressed, via the survey that they benefited from this approach to learn ML concepts. We have also compared the quantum of student learning from the two approaches and found that the approach of delivering machine learning content side-by-side with class concepts provides context, and is therefore, a more effective teaching tool.

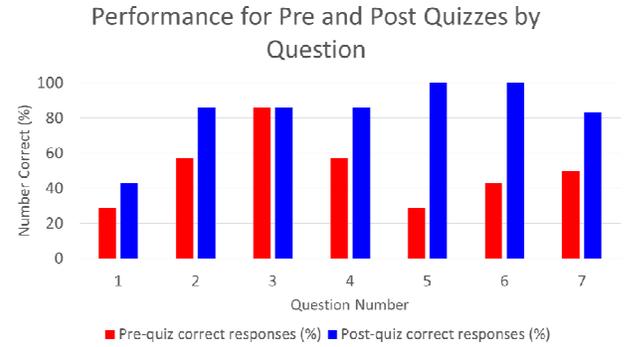

Figure 3. Student responses for the pre- and post-quizzes. The percent correct is shown for all six questions and the overall average for both the pre and post quizzes. Improvement in student performance can be seen in all cases.

The rest of this work-in-progress paper is organized as follows. Section II provides a summary of the stand-alone workshops held. In Section III, we present our approach trialed in the fall semester of 2020. In Section IV, we discuss the assessment results. Finally, in Section V, concluding remarks and future work are presented.

## II. STAND-ALONE MACHINE LEARNING WORKSHOP

The focus of this workshop [4, 5] was to cover the basics of machine learning. Specifically designed datasets (see Figure 2) were created so that key concepts were presented effectively. The workshop was set up in three parts: (1) First, a pre-quiz was administered to determine baseline knowledge; (2) Students were presented machine learning algorithms, both via presentations and hands-on activities; and (3) at the end of the workshop, a post-quiz was administered. The post-quiz contained the same questions as the pre-quiz, along with survey questions. The questions were repeated so that we could gauge if students learned content during the workshop. The survey questions were included to elicit feedback from students to evaluate the effectiveness of the workshop.

The topics covered in the workshop included:

(1) Introduction of multiple machine learning techniques such as k-nearest neighbors (kNN) and support vector machines (SVM).
(2) Comparison between these methods.
(3) Need for a kernel SVM classifier
(4) Principal component analysis (PCA)
(5) Supervised vs unsupervised algorithms

Two survey questions were asked:

(1) "This workshop was fun and easy to follow"; and
(2) "This workshop was helpful and informative".

Students were asked to respond using a one to five Likert-type scale, where 1 indicated "Strongly disagree" and 5 indicated "Strongly agree".

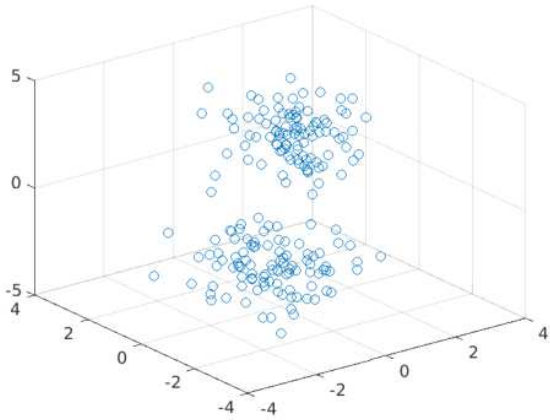

Figure 4. Synthetic dataset with data presented in three dimensions for the project described in Section III-A. Students were asked to visualize the data (to generate the figure shown), and then determine if the data needs to be presented in three dimensions or if could be reduced to fewer dimensions. Using PCA, the data is reduced to one dimension, as shown in Figure 1.

Figure 3 shows the results of the pre-post quiz. Students showed improvement scoring on average 50% on the pre-quiz and then 83.3% on the post-quiz. Student responses were positive, and students showed great interest in the workshop. Average responses on a 1-5 Likert scale for the two survey questions were 4.7 and 4.9, respectively.

## III. IN-CLASS APPROACH

In this section, we present our in-class approach to teach machine learning side-by-side with course content in a Junior-level signals and systems course in our University. In this semester-long effort, we investigated whether machine learning concepts could be introduced effectively this way. Through the semester, various concepts including dimensional analysis and dimensionality reduction, PCA [8-10], classification [11, 12], and digital modulation [13, 14] were introduced to students through a combination of theoretical material and hands-on computer exercises. Topics covered included machine learning theory, PCA, neural networks, and topics in training/testing. In what follows, we will expand on two projects given to students, as well as how they are linked to signals and systems content from the course.

### A. Dimensionality Reduction and PCA

This project introduced basic concepts in machine learning that included classification, dimensionality, and visualization.

*1) Project Summary:* Students were given data that can be visualized in three dimensions (see Figure 4). The dimensionality of the data was investigated using the eigenvalues and eigenvectors of the data. The students were then shown how eigenvalues represent dimensionality, eigenvectors are orthogonal to each other, and how the original data can be reconstructed from the eigenvalues and eigenvalues. Students also use the eigenvalues and eigenvectors of the data to reduce the dimensionality of the data. The data in reduced dimensions was visualized again (see Figure 1), and students were able to see that the main characteristic of the data, its separation into two clear clusters, was retained, despite the processing. Real data, based on keystroke dynamics research [15, 16], was also given to the students, to have them further explore the concepts of visualization and dimensionality reduction.

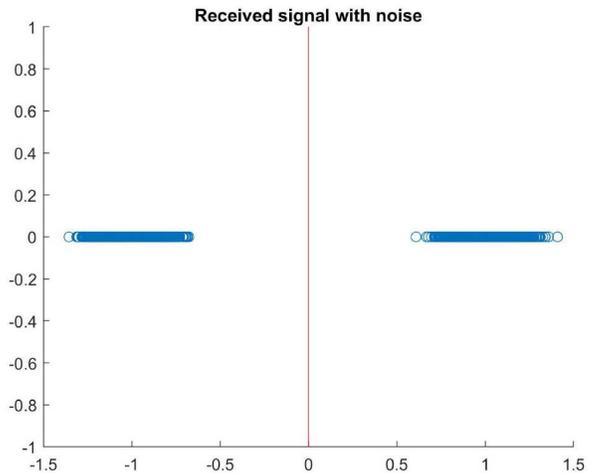

Figure 5. Student work showing a two-class classifier. Two symbols are received in noise, and students design a classifier to distinguish between the two symbols.

*2) Links to signals and systems:* One of the main concepts covered in a signals and systems class is the Fourier series. Fourier series covers concepts related to orthogonality (Fourier basis) and dimensionality (number of harmonics needed to represent signals using the Fourier series). Similar to signal reconstruction, the inverse Fourier series involves the synthesis of time-domain signals from Fourier series coefficients.

### B. Classification

This project looked at separable data, and used machine learning algorithms to classify them into different groups.

*1) Project Summary:* Students work with two simple supervised classifiers. The classifiers were derived (in class) and students implemented them in the project. For both classifiers, students were given "training data" whose classes were known and used to derive the decision boundaries. The algorithms were evaluated on "testing data" that had not been seen by the algorithm. Students were given a digital communications scenario as an application of the classifiers. An example of student work where one of the classifiers is implemented to separate two symbols received in noise is shown in Figure 5.

*2) Links to signals and systems:* In the signals and systems class, students are introduced to the concept of a filter. They study the effects of the systems on signals. The communications system, including the additive noise channel, is modeled as a filter, and students can see the link between the concept of a filter and communications models. In addition to links to signals and systems, the introduction of noise and analysis of errors in classification link the project to concepts covered in other content areas such as probability, statistics, and random variables, providing a crucial multi-disciplinary perspective to the students.

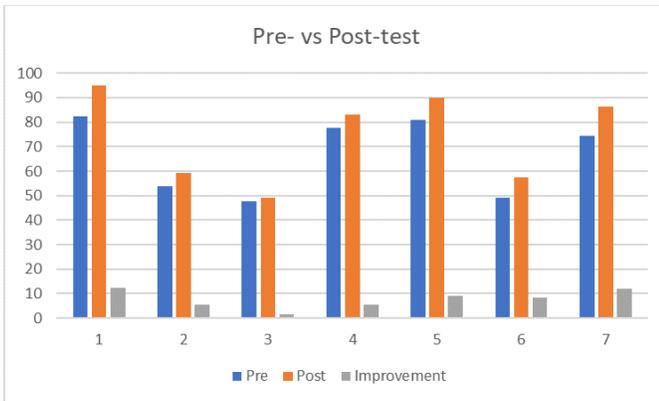

Figure 6. Pre-test vs. post-test. Students showed improvement in all concept areas.

## IV. ASSESSMENT RESULTS

Concept questions were asked in a pre-quiz and post-quiz format, with the pre-quiz at the beginning of the semester and the post-quiz at the end. The questions in the pre- and post-quiz covered concepts including machine learning theory, PCA, neural networks, and topics in training/testing. The expectation was that more students correctly answer the questions in the post-quiz vs. the pre-quiz, and students did show improvement in all concept areas they were exposed to in class (Figure 6).

The post quiz also included three survey questions, and a way to receive long-form feedback from the students, anonymously. About 90% of the students agreed that they liked the activities, learned from them, and felt this side-by-side approach was an effective way for them to learn the material (see Figure 7). Comments in an anonymous survey included "They were a good way to begin dabbling in machine learning applications without us being in over our heads. Additionally, I thought the projects did a good job of introducing the topics while also giving us plenty of room to experiment and learn independently.", "I liked the visualizations", "It helped expand my knowledge on an interesting subject", and "Learning how statistics are applicable from a signals perspective. It was interesting to see how involved stats is within engineering.".

The preceding results show that the proposed method is effective, and students enjoyed the projects. We compared the improvements in student learning from the two approaches. We took the pre- and post- quiz data from the two approaches to determine the average difference in the number of correct questions from the pre- and post- quizzes. We used a Welch's $t$-test [17] to compare the two methods. A $t$-value of 5.7 was obtained. The workshop approach was found to be more effective than the in-class approach at improving the scores from the pre- and post- quizzes at the 5% alpha level ($p<0.001$).

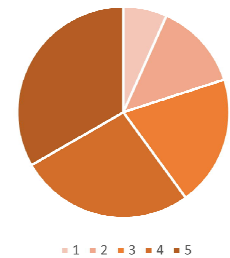

Figure 7. Student response to "I enjoyed working on the projects". More than 90% of the students agree or strongly agree. Students had similar favorable reactions to "The projects were an effective way to learn concepts" and "I learnt some machine learning".

From these results, we can see that while students prefer the side-by-side delivery, workshops appear to be more effective for student learning. Clearly, further investigation is warranted.

## V. CONCLUSIONS AND FUTURE WORK

In this work-in-progress paper, we present our approach to improving undergraduate student understanding of machine learning concepts. While previous approaches have focused on stand-alone workshops, we tried a side-by-side approach in a Junior-level signals and systems course. We hypothesized that providing projects and hands-on activities along with course concepts, provides students links to concepts they are learning, provides reinforcement, and gives context to ML concepts.

We implemented our trial activity with the help of class projects, hands-on activities, and in-class discussions. To evaluate the effectiveness of our approach, we gave students a pre-quiz at the start of the semester and a post-quiz at the end. Students showed improvement across all concepts. Students also found that this approach was effective and that they enjoyed working on the projects. In addition to assessing just this approach, student learning across both approaches, stand-alone workshops and side-by-side content delivery, were compared. Results show while students like the side-by-side delivery better, the workshops showed improved student learning. Clearly, further exploration is warranted.

Based on assessment results and student feedback, future work includes developing methods to seamlessly combine machine learning concepts into the signals and systems course. Also being investigated is the possibility of using this approach in other courses, culminating in a sequence of courses that build student competence. Further evaluation of the effectiveness of this approach is also in progress, including more work in the fall semester of 2021. Assessments will include a more thorough study where instructors have the ability to evaluate if each student meets the set learning objectives for the project. These assessments will be more systematic, with specific rubrics designed to list and assess student outcomes.

## VI. ACKNOWLEGEMENTS

This work is supported in part by the NSF DUE award 1525224 and the NSF CPS award 1646542.

## References

[1] U. S. Shanthamallu, A. Spanias, C. Tepedelenlioglu, and M. Stanley, "A brief survey of machine learning methods and their sensor and iot applications," in 2017 8th International Conference on Information, Intelligence, Systems Applications (IISA), 2017, pp. 1–8.

[2] Dey, Ayon. "Machine learning algorithms: a review." *International Journal of Computer Science and Information Technologies* 7, no. 3 (2016): 1174-1179.

[3] Mohammed, Mohssen, Muhammad Badruddin Khan, and Eihab Bashier Mohammed Bashier. *Machine learning: algorithms and applications*. Crc Press, 2016.

[4] Banavar M.K., Rivera S., Ayotte B., Mack K.V., Barry D., Spanias A., Sahu C. and Yang T. "Teaching Signal Processing Applications using an Android Echolocation App," Computers in Education Journal, vol. 12, no. 1, 2021.

[5] B. Ayotte et al., "Introducing machine learning concepts using hands-on Android-based exercises," 2019 IEEE Frontiers in Education Conference (FIE), Covington, KY, USA, 2019, pp. 1-5.

[6] A. Dixit, U. S. Shanthamallu, A. Spanias, V. Berisha and M. Banavar, "Online Machine Learning Experiments in HTML5," 2018 IEEE Frontiers in Education Conference (FIE), San Jose, CA, USA, 2018, pp. 1-5.

[7] M. K. Banavar, H. Gan, B. Robistow and A. Spanias, "Signal processing and machine learning concepts using the reflections echolocation app," 2017 IEEE Frontiers in Education Conference (FIE), Indianapolis, IN, USA, 2017, pp. 1-5.

[8] Karamizadeh, Sasan, Shahidan M. Abdullah, Azizah A. Manaf, Mazdak Zamani, and Alireza Hooman. "An overview of principal component analysis." *Journal of Signal and Information Processing* 4, no. 3B (2013): 173.

[9] Abdi, Hervé, and Lynne J. Williams. "Principal component analysis." *Wiley interdisciplinary reviews: computational statistics* 2, no. 4 (2010): 433-459.

[10] Wall, Michael E., Andreas Rechtsteiner, and Luis M. Rocha. "Singular value decomposition and principal component analysis." In *A practical approach to microarray data analysis*, pp. 91-109. Springer, Boston, MA, 2003.

[11] Kotsiantis, Sotiris B., I. Zaharakis, and P. Pintelas. "Supervised machine learning: A review of classification techniques." *Emerging artificial intelligence applications in computer engineering* 160, no. 1 (2007): 3-24.

[12] Kotsiantis, Sotiris B., Ioannis D. Zaharakis, and Panayiotis E. Pintelas. "Machine learning: a review of classification and combining techniques." *Artificial Intelligence Review* 26, no. 3 (2006): 159-190.

[13] Chuah, Chen-Nee, David N. C. Tse, Joseph M. Kahn, and Reinaldo A. Valenzuela. "Capacity scaling in MIMO wireless systems under correlated fading." *IEEE Transactions on Information theory* 48, no. 3 (2002): 637-650.

[14] Yang, Hong-Chuan, and Mohamed-Slim Alouini. Order statistics in wireless communications: diversity, adaptation, and scheduling in MIMO and OFDM systems. Cambridge University Press, 2011.

[15] Zhong, Yu, Yunbin Deng, and Anil K. Jain. "Keystroke dynamics for user authentication." In *2012 IEEE computer society conference on computer vision and pattern recognition workshops*, pp. 117-123. IEEE, 2012.

[16] Sahu, Chinmay, Mahesh Banavar, and Jie Sun. "A novel distance-based algorithm for multi-user classification in keystroke dynamics." In *2020 54th Asilomar Conference on Signals, Systems, and Computers*, IEEE, 2020.

[17] Hayter, Anthony J. *Probability and statistics for engineers and scientists.* Cengage Learning, 2012.